\documentclass[letterpaper]{article} 
\usepackage{aaai25}  
\usepackage{times}  
\usepackage{helvet}  
\usepackage{courier}  
\usepackage[hyphens]{url}  
\usepackage{graphicx} 
\usepackage{natbib}  
\usepackage{caption} 
\usepackage{algorithm}
\usepackage{algorithmic}
\usepackage{subcaption}

\usepackage{xcolor}  
\usepackage{amsmath}
\usepackage{amsfonts}

\usepackage{newfloat}
\usepackage{listings}

\usepackage{microtype}
\DeclareCaptionStyle{ruled}{labelfont=normalfont,labelsep=colon,strut=off} 
\floatstyle{ruled}
\newfloat{listing}{tb}{lst}{}
\floatname{listing}{Listing}
\title{
DODT: Enhanced Online Decision Transformer Learning through Dreamer's Actor-Critic Trajectory Forecasting}

\author{
    Eric Hanchen Jiang\textsuperscript{\rm 1}\thanks{Corresponding to: ericjiang0318@g.ucla.edu},
    Zhi Zhang\textsuperscript{\rm 1},
    Dinghuai Zhang\textsuperscript{\rm 2},
    Andrew Lizarraga\textsuperscript{\rm 1},
    Chenheng Xu\textsuperscript{\rm 1},
    Yasi Zhang\textsuperscript{\rm 1},
    Siyan Zhao\textsuperscript{\rm 1},
    Zhengjie Xu\textsuperscript{\rm 3},
    Peiyu Yu\textsuperscript{\rm 1},
    Yuer Tang\textsuperscript{\rm 1},
    Deqian Kong\textsuperscript{\rm 1},
    Ying Nian Wu\textsuperscript{\rm 1}
}
\affiliations {
    \textsuperscript{\rm 1}University of California, Los Angeles, 
    \textsuperscript{\rm 2}Mila, 
    \textsuperscript{\rm 3}University of Michigan, Ann Arbor
}

\begin{document}

\maketitle

\begin{abstract}
Advancements in reinforcement learning have led to the development of sophisticated models capable of learning complex decision-making tasks. However, efficiently integrating world models with decision transformers remains a challenge. In this paper, we introduce a novel approach that combines the Dreamer algorithm's ability to generate anticipatory trajectories with the adaptive learning strengths of the Online Decision Transformer. Our methodology enables parallel training where Dreamer-produced trajectories enhance the contextual decision-making of the transformer, creating a bidirectional enhancement loop. We empirically demonstrate the efficacy of our approach on a suite of challenging benchmarks, achieving notable improvements in sample efficiency and reward maximization over existing methods. Our results indicate that the proposed integrated framework not only accelerates learning but also showcases robustness in diverse and dynamic scenarios, marking a significant step forward in model-based reinforcement learning. 
\end{abstract}

%


\section{Introduction}

Given the recent success of transformer architectures \cite{vaswani2017}, the general framework of the Decision Transformer (DT) is designed for rapid adaptation and enhanced computational rewards by leveraging pre-training data in an offline setting \cite{chen2021}. Building upon the Decision Transformer, the Online Decision Transformer (ODT) is tailored for online reinforcement learning (RL)  settings where decisions must be made in real-time based on streaming data \cite{zheng2022}, while simultaneously learning from the dataset \cite{fu2021}. The key innovation of the ODT lies in its ability to continuously integrate new experiences and dynamically update the policy as new data arrives. This capability is crucial in non-stationary environments where the underlying dynamics may change over time, necessitating timely policy adaptations.

At the core of the Decision Transformer architecture, the decision transformer maintains a replay buffer of recent experiences, utilizing trajectories of states, actions, and rewards \cite{janner2021}. This buffer is used to fine-tune the policy network at regular intervals, ensuring that the decision-making strategy remains aligned with the most recent data. This process provides the agent with the capacity to effectively respond to evolving situations \cite{li2023}. By combining the transformer’s ability to process sequences with online learning, the ODT facilitates a more robust and adaptive approach to decision-making in dynamic environments \cite{bhargava2024}. 

\begin{figure}[b]
\label{fig:1}
\centering
\includegraphics[width=0.47\textwidth]{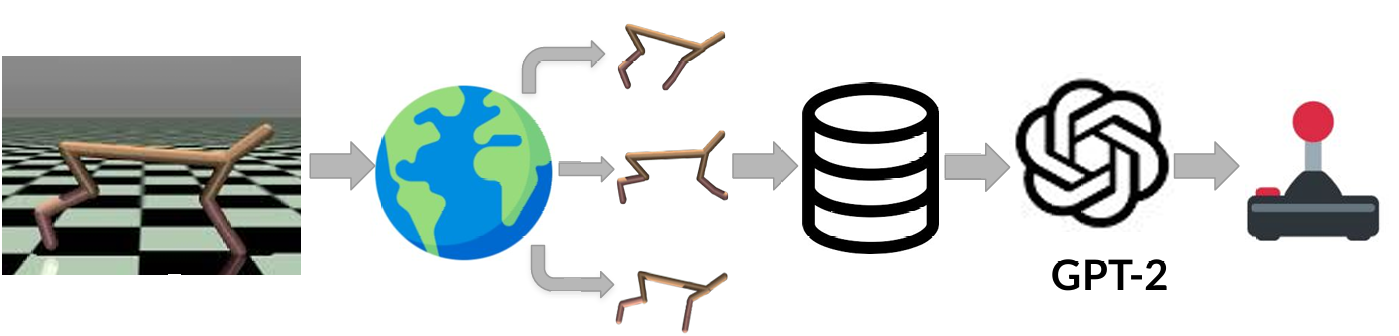} 
\caption{The figure illustrates how the Dreamer's world model infers latent states from environmental observations and stores these trajectories in the replay buffer. Subsequently, the Decision Transformer, fine-tuned on a GPT-2 model, learns from these trajectories to determine the agent's actions. }
\label{fig2}
\end{figure}

Similarly, the Dreamer algorithm \cite{hafner2020}, another popular reinforcement learning approach, utilizes world models to ``dream'' or simulate future states. This capability allows the agent to anticipate the outcomes of its actions without direct interaction with the actual environment, thereby enhancing the efficiency of the learning process by reducing the need for extensive real-world data \cite{hafner2024}. Dreamer operates by learning a latent dynamics model of the environment \cite{hafner2019}, which captures transition probabilities and reward functions. Once trained, this model can generate synthetic trajectories of states, actions, and rewards that the agent can use to improve its policy. The fundamental principle is that by learning in the space of latent representations, the algorithm can perform planning and credit assignment more effectively, even in high-dimensional state spaces \cite{brunnbauer2022}. By imagining outcomes and backpropagating the future rewards, Dreamer optimizes the policy to maximize expected returns. This not only conserves resources but also enables safer training, as the agent can explore various strategies in a simulated environment before executing them in the real world. Consequently, Dreamer is capable of developing sophisticated behaviors even in complex environments with sparse rewards \cite{moerland2022}.

In conventional reinforcement learning problems, models often oscillate between exploiting historical data and exploring new strategies through trial and error. The Dreamer algorithm, excels in simulating potential future states, offering a vision of probable outcomes based on current decisions. Conversely, the Online Decision Transformer capitalizes on accumulated experiences, effectively summarizing past actions and outcomes to optimize immediate decisions. By merging these two methodologies, our framework enables higher reward learning. To this end, we proposed a novel algorithm: \textsl{ \underline{D}ream-to-Control-for-\underline{O}nline-\underline{D}ecision-\underline{T}ransformer (DODT)}. This novel framework uses the base framework of Online Decision Transformer and through a paralleled trained dreamer, the transfer of enhanced trajectory from dreamer to ODT can benefit the overall model. Through numerous experiments, DODT can utilizes the success of ODT and Dreamer, achieving a higher reward.


  We conclude our \textbf{contributions} from three perspectives:
\begin{enumerate}
  \item \textbf{Parallel Training Architecture:} We present the first and novel parallel training methodology that simultaneously leverages the Dreamer model's trajectory generation and the Online Decision Transformer's adaptive learning capabilities, providing a symbiotic framework for decision-making.
  \item \textbf{Trajectory-Informed Decision Making:} Our integration uniquely enables the Online Decision Transformer to be informed by high-fidelity trajectories from the Dreamer, thus enhancing its contextual understanding and response strategies in complex environments.
  \item \textbf{Cross-Model Feedback Mechanism:} We introduce a feedback loop between the Dreamer and the Online Decision Transformer. Our integrated approach demonstrates superior performance across a variety of challenging benchmarks, surpassing traditional methods in terms of sample efficiency and reward maximization.
\end{enumerate}

\section{Related Works} 

\textbf{Decision Transformer: } Recent advancements in Decision Transformers have significantly expanded their capabilities and applications in reinforcement learning. A bootstrapping method was introduced to augment data generation for both online and offline Decision Transformers, enhancing training datasets significantly \cite{wang2022}. Additionally, innovative probabilistic learning objectives and max-entropy sequence modeling have been integrated to balance exploration and exploitation dynamically, addressing the demands of online reinforcement learning environments for decision transformers \cite{ma2023}. Further enhancements include a hierarchical decision-making structure, where high-level policies generate prompts that guide low-level action generation, improving decision granularity \cite{moerland2022}, and the combination of trajectory modeling with value-based methods, which aligns specified target returns with expected action returns to boost performance in stochastic settings \cite{wang2023b}). Additionally, leveraging latent diffusion models for optimizing suboptimal trajectory portions from static datasets \cite{Venkatraman2022} and employing robust planning frameworks that treat planning as latent variable inference have further enhanced the long-term decision-making capabilities of Decision Transformers \cite{kong2024}. 

\textbf{Dreamer: } At the same time, the Dreamer have been enhanced through various innovative approaches as well. The Dreamer model has been extensively advanced by integrating transformers to enhance the deterministic state prediction from observations  \cite{zeng2022}. Transitioning from recurrent neural networks to transformer networks within the world model has significantly improved the efficiency of state predictions  \cite{chen2022, ding2024}. The adaptation of Dreamer for multi-task reinforcement learning uses diffusion models to optimize offline decision-making \cite{he2023}. Extensions to the Dreamer framework allow handling of diverse tasks through world models that predict future states and rewards from abstract representations \cite{hafner2024}. Furthermore, the use of prototypical representations instead of high-dimensional observation reconstructions \cite{deng2021}, along with conditional diffusion models for long-horizon predictions \cite{zhao2023}, and enhancement of exploration using latent state marginalization \cite{zhang2023}, collectively push the boundaries of model-based RL.

\textbf{Teacher to Student Model: } The Student to Teacher model leverages the dynamics of guided learning to enhance the efficiency and scalability of reinforcement learning systems. The TGRL algorithm integrates the teacher-student learning framework with reinforcement learning, facilitating enriched policy learning experiences \cite{shenfeld2024}. Curriculum learning approaches have also been significant, framing task sequencing within a meta Markov Decision Process to systematically improve sample efficiency \cite{schraner2022}. The application of large language models as teachers to guide smaller, specialized student agents offers a novel approach to scaling down complex decision processes \cite{zhou2024}. Furthermore, advancements in multi-agent systems, where experiences are shared between agents, enhance collective learning capabilities, demonstrating improved scalability and efficiency \cite{wang2023}. \\

\begin{figure*}[t]
\centering
\includegraphics[width=1\textwidth]{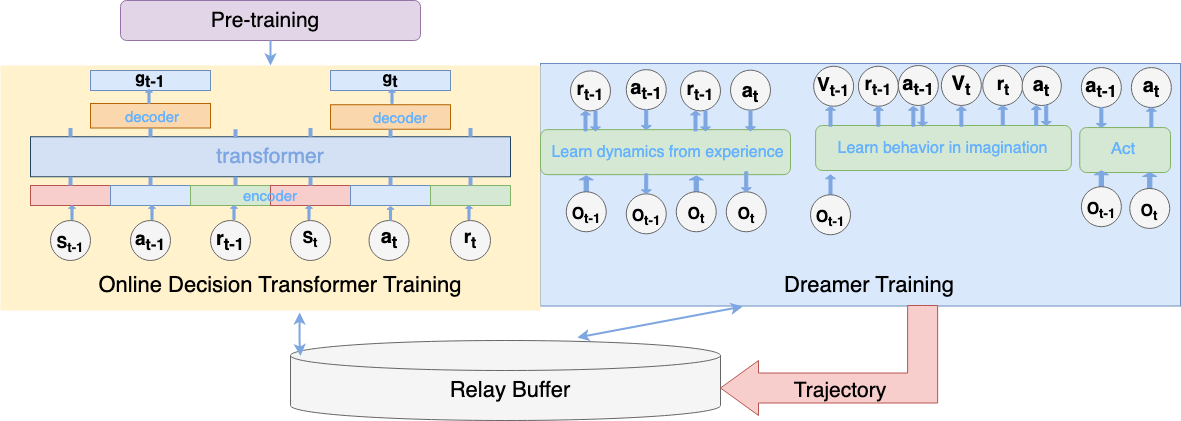} %
\caption{ This diagram illustrates the combined training approach of the Online Decision Transformer (left) and the Dreamer model (right). The Online Decision Transformer refines decision-making strategies using historical data and ongoing interactions stored in the relay buffer. Concurrently, the Dreamer model projects future trajectories, enriching the relay buffer with simulated experiences that enhance the predictive capabilities of the system. This integrated framework allows for dynamic adaptation and improved decision-making in complex environments.}
\end{figure*}

\section{Preliminaries}
\subsection{Online Decision Transformer}

\begin{algorithm}[tb]
\caption{Online Decision Transformer (ODT)}
\label{alg:odt}
\textbf{Input}: Offline data $T_{\text{offline}}$, rounds $R$, exploration RTG  (Reward To Go)$T_{\text{online}}$, buffer size $N$, gradient iterations $I$, pre-trained policy $\pi_\theta$ \\
\textbf{Initialization}: Replay buffer $T_{\text{replay}} \leftarrow$ top $N$ trajectories in $T_{\text{offline}}$
\begin{algorithmic}[1] 
\FOR{$\text{round} = 1, \ldots, R$}
    \STATE Trajectory $\tau \leftarrow$ Using $M$ and $\pi_\theta(\cdot|s, g(T_{\text{online}}))$
    \STATE $T_{\text{replay}} \leftarrow (T_{\text{replay}} \setminus \{\text{oldest trajectory}\}) \cup \{\tau\}$
    \STATE $\pi_\theta \leftarrow$ Finetune ODT on $T_{\text{replay}}$ for $I$ iterations using Training Main Loop
\ENDFOR
\end{algorithmic}
\end{algorithm}

Online Decision Transformer (ODT) represents a significant advance in the application of transformers to reinforcement learning (RL). It extends the Decision Transformer (DT) architecture to online settings, adapting the transformer architecture for dynamic environments and real-time decision-making tasks. This adaptation is crucial for RL applications where an agent must continuously learn and adapt based on new data while interacting with an environment. In traditional reinforcement learning, decision-making often relies on policies learned from historical data or through iterative interactions with an environment. These methods can be inefficient and slow to adapt to changes in dynamic scenarios. The ODT framework addresses these challenges by leveraging the sequential processing capabilities of transformers to model policies based on both past and current interactions, integrating learning and decision-making in an online fashion.

The core of ODT is a transformer architecture trained to optimize a sequence modeling objective that predicts the next action based on a history of states, actions, and rewards. Given a history encoded as sequences, the ODT models the conditional probability of actions given past experiences, formulated as:
\[
\pi(a_t | s_t, g_t) \approx P(a_t | \text{context}),
\]
where \( g_t \) represents the return-to-go, a sum of future rewards, and \( s_t \) denotes the current state. The context comprises past states, actions, and achieved rewards up to time \( t \).

The policy is refined using a replay buffer \( T_{\text{replay}} \) that stores trajectories:
\[
T_{\text{replay}} = \{\tau_1, \tau_2, \dots, \tau_N\},
\]
where each \( \tau_i \) is a trajectory containing sequences of states, actions, and rewards. During training, this buffer is continuously updated by replacing the oldest trajectories with new ones obtained from recent environment interactions, ensuring that the policy adapates to the most recent data.

The ODT utilizes the transformer's capability to process sequences of data to dynamically update its policy based on the replay buffer. The policy \( \pi_\theta \) is optimized by fine-tuning the transformer model on sequences drawn from \( T_{\text{replay}} \), using the objective:
\[
\pi_\theta^* \leftarrow \arg\max_{\pi_\theta} \mathbb{E}\left[ \sum_{t=0}^{T} \gamma^t r_t \right],
\]

where \( r_t \) is the reward at time \( t \) and \( \gamma \) is the discount factor.

During online interactions, the ODT first collect data from the environment using the current policy \( \pi_\theta \) then update the replay buffer \( T_{\text{replay}} \) by incorporating new trajectories and discarding the oldest. After that, it refines the policy \( \pi_\theta \) by training on a sampled batch from \( T_{\text{replay}} \). This continuous loop of feedback and adaptation allows the ODT to maintain a policy that is responsive to the evolving dynamics of the environment. The integration of a transformer-based sequence model with an RL policy training framework enables the ODT to leverage the strengths of both sequence modeling and reinforcement learning techniques.

\subsection{Dreamer}

        

\begin{algorithm}[tb]
\caption{Dreamer}
\label{alg:dreamer}
\textbf{Input}: Number of data sequences $L$, Imagine trajectory length $H$, Dynamics learning steps $C$, Environment interaction steps $T$ \\
\textbf{Parameter}: Initial random seed episodes $S$, Neural network parameters $\theta, \phi, \psi$ initialized randomly \\
\textbf{Initialization}: Dataset $D$ with $S$ random seed episodes and $B$ batch size, and initialize $k$ starting index for each sequence in the data batch.  
\begin{algorithmic}[1] 
\WHILE{not converged} 
    \FOR{update step $c = 1 \ldots C$}
        \STATE // Dynamics learning
        \STATE Draw $B$ data sequences $\{(\mathbf{a}_t, \mathbf{o}_t, r_t)\}_{t=k}^{k+L} \sim D$
        \STATE Initialize $s_{t-1}$ from last state of previous batch if available, else initialize randomly
        \STATE Compute model states $s_t \sim p_\zeta(s_t | s_{t-1}, a_{t-1}, o_t)$  
        \STATE Update $\zeta$ using representation learning 
        
        \STATE // Behavior learning
        \STATE Imagine trajectories $\{(s_r, a_r)\}_{r=t}^{t+H}$ from each $s_t$
        \STATE Predict rewards $E(r_t | s_t)$ and values $V(s_t)$
        \STATE Compute value estimates $V_{\psi}(s_t)$ 
        \STATE Compute value estimates $v_{\psi}(s_t)$ of $V(s_t)$ and prediction $r_t$ of $E(r_t | s_t)$. 
        \STATE Update $\phi \gets \phi + \alpha \nabla_\phi \sum_{r=t}^{t+H} v_{\psi}(s_r)$
        \STATE Update $\psi \gets \psi - \alpha \nabla_\psi \frac{1}{2} \sum_{r=t}^{t+H} (v_{\psi}(s_{r+1}) + \gamma r_r - v_{\psi}(s_r))^2$.
    \ENDFOR
    
    \STATE // Environment interaction
    \STATE $\mathbf{o}_1 \gets \text{env.reset}()$
    \FOR{time step $t = 1 \ldots T$}
        \STATE Compute $s_t \sim p_\zeta(s_t | s_{t-1}, a_{t-1}, o_t)$
        \STATE Compute $a_t \sim q_\phi(a_t | s_t)$ with the action model
        \STATE Add exploration noise to $a_t$
        \STATE Execute action $a_t$ and observe reward $r_t$ and new observation $\mathbf{o}_{t+1}$
        \STATE Add experience to $D \gets D \cup \{(\mathbf{o}_t, a_t, r_t)\}_{t=1}^{T}$
    \ENDFOR
\ENDWHILE
\end{algorithmic}
\end{algorithm}

The Dreamer algorithm represents a significant step forward in latent dynamics learning for control by leveraging model based reinforcement learning, mostly for model based RL. By abstracting the observation space into a compact latent space, the dreamer can efficiently predicts future states and rewards, enabling it to plan and learn policies entirely through latent imagination.

The world model in Dreamer consists of three key components:
\begin{itemize}
    \item \textbf{Representation Model}: $p_\zeta(s_t | s_{t-1}, a_{t-1}, o_t)$, which encodes observations into a latent state, integrating past actions and states.
    \item \textbf{Transition Model}: $q(s_t | s_{t-1}, a_{t-1})$, which predicts the next latent state given the current state and action, facilitating the generation of future trajectories without real-world interaction.
    \item \textbf{Reward Model}: $q_{\xi}(r_t | s_t)$, which estimates the immediate reward from the current latent state, crucial for evaluating the desirability of states within imagined trajectories.
\end{itemize}

Dreamer utilizes latent imagination to learn optimal behaviors by simulating trajectories in the latent space, derived from the learned world model. This approach allows Dreamer to perform efficient, farsighted planning by propagating value estimates backward through imagined trajectories. The key mathematical formulations in this process include the \textbf{Action Model}: $a_\tau \sim q_\phi(a_\tau | s_\tau)$, which optimizes actions to maximize expected returns, and the \textbf{Value Model}: $v_\psi(s_\tau) \approx \mathbb{E}\left[\sum_{\tau=t}^{t+H} \gamma^{\tau-t} r_\tau | s_\tau\right]$, which estimates the value of latent states over a finite horizon $H$.

The optimization objectives are:
\begin{itemize}
    \item \textbf{Action Optimization}: 
    \[
    \max_\phi \mathbb{E}_{p_\zeta, q_\phi}\left[\sum_{\tau=t}^{t+H} v_\psi(s_\tau)\right],
    \] 
    aiming to find the policy parameters $\phi$ that maximize the sum of discounted future values estimated by the value model.
    \item \textbf{Value Regression}:
     \[
    \min_\psi \mathbb{E}_{p_\zeta, q_\phi}\left[\frac{1}{2} \sum_{\tau=t}^{t+H} \|\gamma v_\psi(s_{\tau+1}) + r_{\tau} - v_\psi(s_\tau)\|^2\right],
    \] 
    minimizing the prediction error of the value model, aligning it with the computed value estimates to ensure consistency and stability in policy evaluation.
\end{itemize}

Dreamer's integration of deep learning with latent variable models for reinforcement learning showcases several advantages over both traditional model-based and model-free methods. By optimizing behavior in a compact, learned representation of the world, Dreamer achieves remarkable data efficiency and scalability, effectively handling environments with complex, high-dimensional sensory inputs. This makes it a powerful tool for a wide range of applications, from robotics to virtual simulations, where sample efficiency and rapid adaptation to new scenarios are critical.

\section{Algorithm: Dreamer Online Decision Transformer for RL}

\begin{figure*}[t]
\centering
\begin{subfigure}{0.5\columnwidth}
  \centering
  \includegraphics[width=\linewidth]{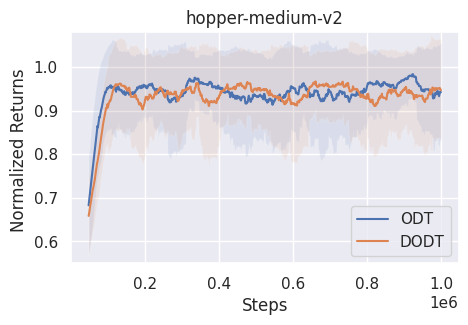}
  \caption{Hopper-v2}
\end{subfigure}
\begin{subfigure}{0.5\columnwidth}
  \centering
  \includegraphics[width=\linewidth]{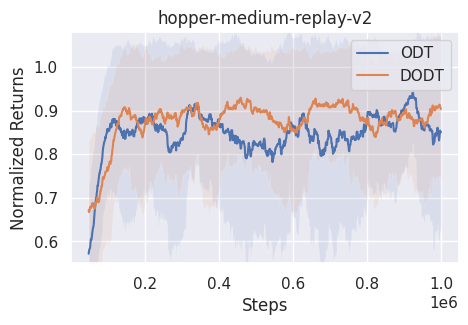}
  \caption{Hopper-v2 Replay}
\end{subfigure}
\begin{subfigure}{0.5\columnwidth}
  \centering
  \includegraphics[width=\linewidth]{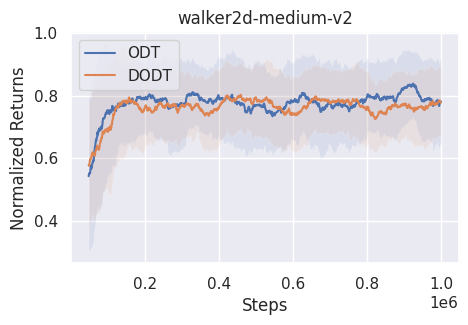}
  \caption{Walker2d-v2}
\end{subfigure}
\begin{subfigure}{0.5\columnwidth}
  \centering
  \includegraphics[width=\linewidth]{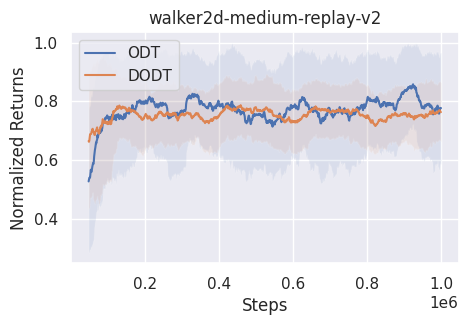}
  \caption{Walker2d-v2 Replay}
\end{subfigure}
\begin{subfigure}{0.5\columnwidth}
  \centering
  \includegraphics[width=\linewidth]{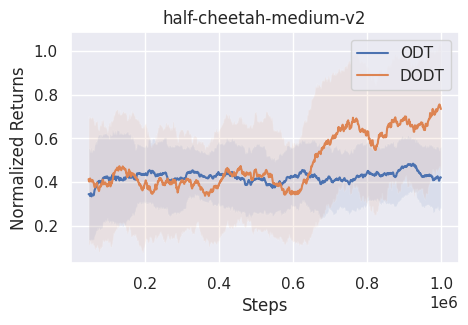}
  \caption{HalfCheetah-v2}
\end{subfigure}
\begin{subfigure}{0.5\columnwidth}
  \centering
  \includegraphics[width=\linewidth]{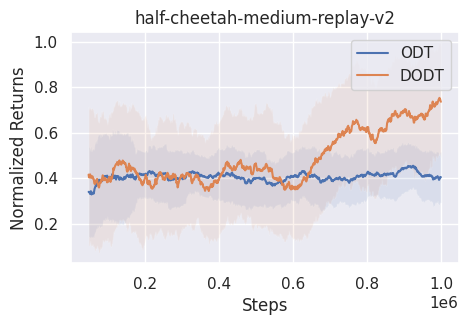}
  \caption{HalfCheetah-v2 Replay}
\end{subfigure}
\begin{subfigure}{0.5\columnwidth}
  \centering
  \includegraphics[width=\linewidth]{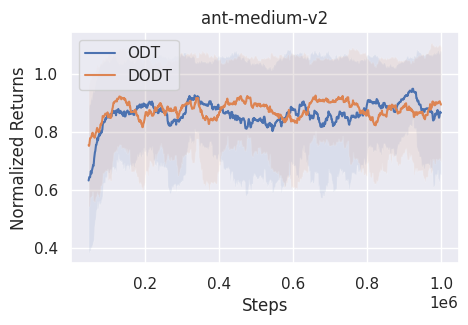}
  \caption{Ant-v2}
\end{subfigure}
\begin{subfigure}{0.5\columnwidth}
  \centering
  \includegraphics[width=\linewidth]{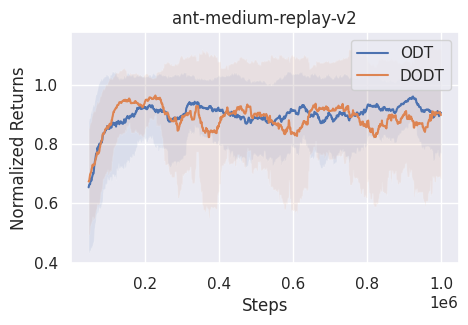}
  \caption{Ant-v2 Replay}
\end{subfigure}
\caption{This figure shows the convergence graphs of Online Decision Transformer (ODT) and Dreamer Online Decision Transformer (DODT) in various MuJoCo environments. The graphs depict normalized returns over steps for different environments: (a) Hopper-v2 to (h) Ant-v2 Replay.}
\end{figure*}


Our proposed algorithm (DODTS, Algorithm \ref{alg:dodt}) integrates the Online Decision Transformer (ODT, Algorithm 1) and Dreamer (Algorithm 2) into a cohesive framework to enhance learning in complex environments. This integration exploits the generative model capabilities of Dreamer and the decision-making prowess of ODT, providing a robust solution to decision-making tasks in dynamic environments.

In the initialization phase, The DODT begins by establishing separate replay buffers: \(T_{\text{replay}}\) for ODT and \(D\) for Dreamer, each tailored to their respective learning needs. Model parameters—\(\pi_\theta\) for ODT and \(\phi, \psi\) for Dreamer—are initialized using either pre-trained states to accelerate learning or random weights to ensure exploratory diversity from the outset.

As the training progresses, Dreamer actively interacts with the environment to generate new trajectories. These trajectories are not mere sequences of actions and outcomes; they are imbued with predictive insights that enhance Dreamer’s generative and predictive capabilities. Concurrently, ODT generates its own trajectories based on the current policy \(\pi_\theta\), each optimized towards a specified reward-to-go. These trajectories are then systematically incorporated into \(T_{\text{replay}}\), ensuring a continually updated dataset that enriches the learning phase. 

\begin{algorithm}[p]
\caption{DODT: Parallel ODT Training with Dreamer Trajectories}
\label{alg:dodt}
\textbf{Input}: {Offline data $T_{\text{offline}}$, exploration RTG $T_{\text{online}}$, buffer sizes $N$, $D$, training rounds $R$, gradient iterations $I$.} \\
\textbf{Initialization}: Initialize replay buffers $T_{\text{replay}}$ and $D$ as per Algorithms 1 and 2, Initialize policies $\pi_\theta$ (ODT) and $\phi, \psi$ (Dreamer) with pre-training or random weights, Load environment and set up necessary configurations
\begin{algorithmic}[1] 
\FOR{$\text{round} = 1, \ldots, R$}
    \STATE \textit{// Dreamer Interaction Phase (Algorithm 2)}
    \STATE $\mathbf{o}_1 \gets \text{env.reset}()$
    \FOR{time step $t = 1 \ldots T$}
        \STATE Use Dreamer to compute $a_t \sim q_\phi(a_t | s_t)$ for the current state
        \STATE Execute $a_t$ in the environment, observe new state $\mathbf{o}_{t+1}$, reward $r_t$
        \STATE Update Dreamer's dataset $D \gets D \cup \{(\mathbf{o}_t, a_t, r_t)\}$
        \STATE Use Dreamer's model to perform learning updates
    \ENDFOR
    
    \STATE \textit{// ODT Interaction Phase (Algorithm 1)}
    \STATE $\tau \leftarrow$ Generate trajectory using Dreamer's $\pi_\theta$ for exploration with RTG $T_{\text{online}}$
    \STATE Update ODT's replay buffer $T_{\text{replay}} \leftarrow (T_{\text{replay}} \setminus \{\text{lowest reward trajectory}\}) \cup \{\tau\}$
    \STATE Finetune $\pi_\theta$ using ODT on $T_{\text{replay}}$ for $I$ gradient iterations
    
    \STATE \textit{// Evaluate Performance}
    \STATE Evaluate the combined performance of Dreamer and ODT
    \STATE Log performance metrics
\ENDFOR
\end{algorithmic}
\end{algorithm}

The integration of Dreamer’s foresighted trajectories allows ODT to preemptively assess the consequences of potential decisions, enhancing its strategic decision-making process. The policy \(\pi_\theta\) undergoes fine-tuning by assimilating these insights, thus aligning more closely with both immediate environmental feedback and predictive foresights provided by Dreamer. This calibration enhances the decision-making accuracy and adaptability of ODT.

Each training round deepens the collaborative interaction between Dreamer and ODT, significantly boosting the decision transformer’s capabilities and sample efficiency. This mutual enhancement facilitates a broader experiential learning base, enabling ODT to foresee and strategically respond to varied scenarios. The models’ continuous information exchange and iterative updates through gradient iterations and buffer management maximize their predictive and decision-making performance.

Consequently, this innovative approach not only accelerates the learning process but also enhances the decision-making quality and predictive accuracy of the system. By leveraging the strengths of both models, DODT adeptly addresses the complexities inherent in dynamic tasks, offering a significant advancement in the field of reinforcement learning.



\section{Experiments}

\begin{table*}[ht]   
\centering  
\small
\begin{tabular}{c|| c c c c | c c} 
 \hline
 Dataset & DT & QDT & ODT (Offline) & DODT (Offline) & ODT & DODT \\ [0.5ex] 
 \hline
 Hopper - medium & 61.03 ± 5.11 & 66.5 ± 6.3 & 66.59 ± 3.26 & 73.84 ± 3.68 & \textbf{97.94 ± 2.10} & 96.84 ± 2.19 \\ 
 Hopper - medium -replay & 62.75 ± 15.05 & 52.1 ± 20.3 & 86.64 ± 5.41 & 75.16 ± 5.23 & 88.89 ± 6.33 & \textbf{90.31 ± 5.23} \\ 
 Walker2d - medium & 72.03 ± 4.32 & 67.1 ± 3.2 &72.19 ± 6.49 & 66.53 ± 7.65 & 76.79 ± 2.30 & \textbf{77.13 ± 2.29} \\ 
 Walker2d - medium -replay & 42.53 ± 15.36 & 58.2 ± 5.1 & 68.92 ± 4.79 & 67.59 ± 5.43 & 76.86 ± 4.04 & \textbf{77.61 ± 3.39} \\  
 Half-cheetah - medium & 42.43 ± 0.30 & 42.3 ± 0.4 & 42.72 ± 0.46 & 63.33 ± 3.84 & 42.16 ± 1.48 & \textbf{60.93 ± 6.83} \\ 
 Half-cheetah - medium -replay & 35.92 ± 1.56 & 35.6 ± 0.5 & 39.99 ± 1.61 & 61.63 ± 7.45 & 40.42 ± 1.61 & \textbf{57.82 ± 5.79} \\ 
 Ant - medium & 93.56 ± 4.94 & / & 91.33 ± 4.13 & 89.82 ± 4.81 & 90.79 ± 5.80 & \textbf{92.01 ± 4.91} \\ 
 Ant - medium -replay & 89.08 ± 5.33 & / & 86.56 ± 3.26 & 90.92 ± 5.93 & 91.57 ± 2.73 & \textbf{93.54 ± 6.31} \\ 
 \hline
 Sum & 499.33  &    & 554.94 & 588.82
 & 605.02 & \textbf{646.19} \\ 
 \hline
 antmaze - umaze & 53.3 ± 5.52 &  57.3 ± 8.2 & 53.1 ± 4.21 & 61.7 ± 4.92 & \textbf{88.5 ± 5.88} & 86.4 ± 5.02 \\ 
 antmaze - umaze-diverse & 52.5 ± 9.89 &  58.6 ± 3.3 & 50.2 ± 6.69 & 57.6 ± 7.21 & 56.00 ± 5.69 & \textbf{74.7 ± 5.16} \\ 
 \hline
 Sum & 105.8 & 115.9 & 103.3 & 123.3 & 144.5 & \textbf{161.1} \\ 
 \hline
\end{tabular}
\caption{Here, we showed the performance comparison of various decision transformer models across multiple datasets and conditions. The table showcases results for Decision Transformer (DT), Q-learning Decision Transformer (QDT), Online Decision Transformer (ODT), and Dreamer Online Decision Transformer (DODT) in both offline and online settings. Bold values denote the highest scores in each category, illustrating the effectiveness of the models under different experimental conditions.}

\label{combined_table}
\end{table*}

We conducted the experiments within the MuJoCo simulation environment \cite{todorov2012}, and a detailed comparative analysis was performed between the Online Decision Transformer (ODT) and the Dreamer Online Decision Transformer (DODT). Both of these models were evaluated across a suite of tasks designed to probe their efficacy under varying conditions reflective of real-world complexity. During training, we noted that ODT and DODT consumed 22.1 GB and 24.6 GB of RAM, respectively, and the experiments ran for 10.5 hours and 12.3 hours per experiment. All experiments were conducted on one A100 graphics card.

By combining trajectories, we observed that in the replay buffer, which maintains a maximum of 1000 trajectories where the least beneficial are discarded each round, there were between 203 to 717 Dreamer trajectories added, with the most significant addition observed in the Half-Cheetah experiment. This strategic management of the replay buffer played a crucial role in optimizing learning outcomes and enhancing model performance.

\begin{figure}[H]
\centering
\includegraphics[width=0.95\columnwidth]{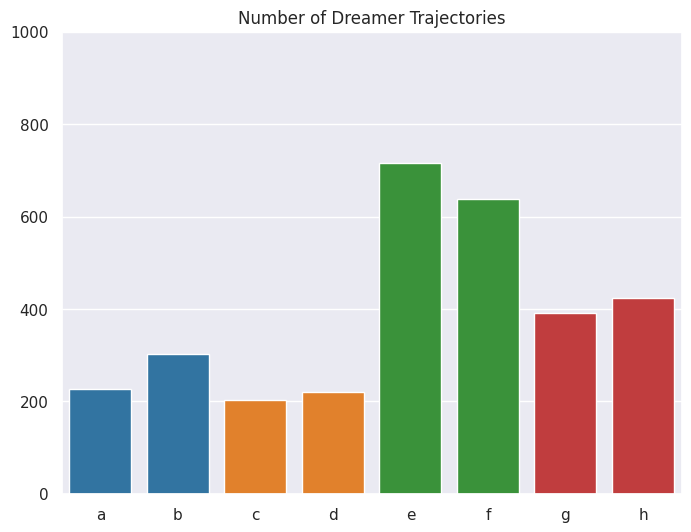}
\caption{This figure shows the number of benefited trajectories used from Dreamer to the Online Decision Transformer, which aided in achieving higher rewards for environments (a) Hopper-v2 to (h) Ant-v2 Replay.}
\label{fig1}
\end{figure}

We analyzed the performance of the Online Decision Transformer (ODT) and the Dreamer Online Decision Transformer (DODT) across various tasks. Results indicate that the Dreamer Online Decision Transformer (DODT) consistently outperformed the standalone Online Decision Transformer (ODT) across various reinforcement learning environments. Notably, in the Hopper and Half-Cheetah environments, DODT demonstrated substantial improvements in both offline and online configurations. For instance, in the Half-Cheetah-medium setting, DODT achieved a score of 60.93, significantly surpassing ODT's 42.16. This highlights DODT's enhanced ability to handle dynamic tasks with complex, rapidly changing conditions, owing to its integration of Dreamer's anticipatory trajectory generation. Additionally, the results showed great improvements, observed in the "Antmaze - umaze-diverse" task, where DODT significantly outperformed ODT with a score of 74.7 compared to 56.00. This performance is indicative of DODT's robust adaptability in complex pathfinding and exploration tasks, benefiting from the anticipatory insights provided by Dreamer.

Collectively, these results not only validate the hypothesis that integrating generative trajectory modeling with adaptive decision frameworks enhances reinforcement learning applications but also demonstrate DODT’s effectiveness across a spectrum of tasks with a total score of 646.19, significantly higher than ODT's 605.02. The consistent superior performance of DODT in handling diverse and dynamic conditions suggests it as a promising framework for advancing autonomous decision-making systems in real-world scenarios.

\section{Conclusion and Future Work}

This paper introduced the Dreamer Online Decision Transformer (DODT), an enhanced algorithm that integrates trajectory generation with decision-making, achieving superior performance in the MuJoCo environment through higher rewards, better efficiency, and robustness. The integration of Dreamer’s anticipatory capabilities enables a more proactive and adaptive approach to decision-making, setting a new benchmark in reinforcement learning strategies. Ongoing improvements will enhance the algorithm’s flexibility, making it a viable option for a broader range of applications. Future efforts will focus on reducing computational requirements, minimizing reliance on pre-trained data, enhancing real-world applicability, and expanding scalability to multi-agent systems and various environments.

\appendix

\end{document}